# A New De-blurring Technique for License Plate Images with Robust Length Estimation

P. S. Prashanth Rao
School of Electronics Engineering
Vellore Institute of Technology
Vellore, India
psprao95@gmail.com

Rajesh Kumar Muthu, Senior Member, IEEE
School of Electronics Engineering
Vellore Institute of Technology
Vellore, India
mrajeshkumar@vit.ac.in

*Abstract* — Recognizing a license plate clearly while seeing a surveillance camera snapshot is often important in cases where the troublemaker vehicle(s) have to be identified. In many real world situations, these images are blurred due to fast motion of the vehicle and cannot be recognized by the human eye. For this kind of blurring, the kernel involved can be said to be a linear uniform convolution described by its angle and length. We propose a new de-blurring technique in this paper to parametrically estimate the kernel as accurately as possible with emphasis on the length estimation process. We use a technique which employs Hough transform in estimating the kernel angle. To accurately estimate the kernel length, a novel approach using the cepstral transform is introduced. We compare the de-blurred results obtained using our scheme with those of other recently introduced blind de-blurring techniques. The comparisons corroborate that our scheme can remove a large blur from the image captured by the camera to recover vital semantic information about the license plate.

## I. INTRODUCTION

Very often, having information related to the license plate is important for identifying vehicles involved in traffic violations. Surveillance cameras are available these days for reporting traffic incidents on roads and highways. However, the vehicle motion during the exposure time of the camera blurs the image that it captures.

In the situation where the vehicle is over-speeding, the blurring caused due to vehicle motion is severe and the license plate information in such images can be hardly recovered [2] - [5]. It is hence, important to improve the quality of the snapshot so that the characters in the license plate can be recognized easily to get the required vehicle information.

Several blind image de-blurring (BID) methods have been proposed in the past for solving the blurring problem in images. Some real world cases, still however, pose a challenge to these techniques. Image blurring can be mathematically formulated as

$$P(x, y) = n(x, y) * M(x, y) + Q(x, y) \quad (1)$$

where P is the image degraded by blurring, n is the kernel, M is the original sharp image, Q is the additive white Gaussian noise, and * denotes convolution. In case n is spatially invariant, the problem is a uniform BID problem. Otherwise, it is a non-uniform BID problem.

Many BID algorithms require prior knowledge regarding the kernel to avoid going in the wrong direction while de-blurring the image. A majority of these algorithms estimate the kernel while simultaneously applying a non-blind image de-blurring (NBID) algorithm recursively to get the de-blurred image [6] - [10]. Another way is to accurately determine the kernel first and then apply an NBID algorithm just once to get the final result [11] - [13].

In many real world cases, kernels responsible for blur are parametric in nature. Some examples are blur due to uniform speed and out-of-focus blur [12]. The problem of accurately estimating a blur kernel can be tackled better when seen as a parametric estimation problem. Algorithms for parametric kernel estimation utilise the property that a linear uniform kernel's spectrum is similar to the sinc function which can be distinguished from a natural image [12], [13]. Oliveira et al. [12] assumed natural images to be isotropic, which applies to only large images. For a small image, the spectrum depends on the content of the image such as large scale edge [1].

If the kernel is non-linear in nature, the problem is more complicated since a non-uniform kernel has many degrees of freedom [1]. To simplify kernel estimation in this scenario, non-uniform blurring is assumed to be caused by the projection transform [17] - [19]. Zheng et al. [17] estimated the normal vector of camera scene plane and the camera's direction of motion to handle the blur in both forward and reverse directions. Gupta et al. [18] reduced the camera motion to a 3-D subspace consisting of roll, x- and y- translations. Whyte et al. [19] assumed the blur from a camera shake to be due to the 3-D rotation of the camera, which can be approximated by a roll, yaw and pitch.

Our aim in this paper is to solve the problem of blurring in captured images of over-speeding vehicles which are severely degraded in order to extract information about the license plate. We aim to recover an image in having easily recognizable license plate characters. The kernel is assumed to be a linear motion kernel and its estimation can be viewed as determining the parameters that define it: *angle* and *length*. The angle of the kernel is found by a method employing the Hough transform. A new robust cepstral technique is used for kernel length estimation. Once we find these parameters, we use a simple NBID algorithm to get the final result. The flow chart of our proposed scheme is shown in Figure 1.

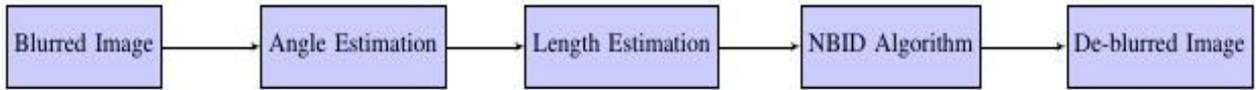

Fig.1. Block diagram of our scheme

The remaining paper consists of the following sections. In Section II, we explain our de-blurring scheme in detail. The results obtained are compared with recent BID techniques and analysed in Section III. We will conclude our paper in section IV.

## II. OUR KERNEL ESTIMATION METHOD

The motion kernel is determined by the relative motion between the vehicle and the surveillance camera during the time of exposure [1]. When the exposure period is small and the vehicle is moving quickly, the motion can be said to be linear and the speed can be assumed as constant. In such cases, the kernel causing the blur can be modeled as a linear uniform motion kernel parametrically described by angle and length [15].

We find the kernel angle by a method employing the Hough transform in section II-A. We then use this angle to find the kernel length in section II-B. We use a new robust cepstral technique for accurately estimating the length. We then use a convenient NBID algorithm to deconvolute our blurred image which is described in section II-C. Figure 2 illustrates an example of a blurred license plate image before and after de-blurring using our scheme.

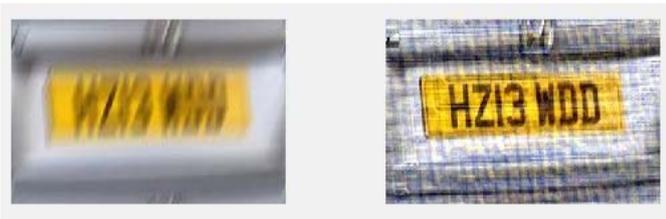

Fig.2. Example of a blurred image and the image after de-blurring using our scheme

*A. Angle Estimation*

To estimate the kernel angle with a good degree of accuracy, we use a technique involving the Hough transform. The Hough transform is applied to detect patterns such as lines, ellipses and circles in an image [15]. The polar form of representing a line is given by

$$\rho = x \cos \theta + y \sin \theta \quad (2)$$

where (x, y) are the Cartesian coordinates of a point on the line, is the angle between the x-axis and the perpendicular from the origin to the line and ρ is the length of the perpendicular. In other words, a line can be described by a pair of coordinates (ρ, θ) in the polar domain. The Hough transform, thus, helps us in detecting lines as points in the Hough domain. Figure 3 illustrates mapping from image domain to the Hough domain.

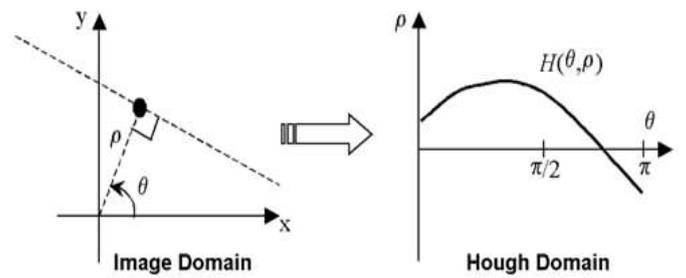

Fig.3. Hough transformation

The Hough transform divides the parameter space into an accumulator array. For each point $(x_j, y_j)$ in the image, the curve given by

$$v = x_j \cos \theta + y_j \sin \theta \quad (3)$$

is recorded in the accumulator array (θ is in radians). Each element of the accumulator array counts the number of curves passing through it. When all image points are recorded in the array, we search it for elements having a high count. If the count in an element (v, θ) is N, then N image points are lying on the line whose Hough domain parameters are (v, θ).

We find the cepstrum for each channel of the blurred image separately. The cepstral transform is defined as:

$$C [ g (x, y) ] = F^{-1} [ \log | F (g (x, y)) | ] \quad (4)$$

We obtain the edge map of the cepstrum. For each point $(x_p, y_p)$ in it, we perform the following computation:

for $\theta = \theta_{min}$ to $\theta_{max}$

$$v = x_p \cos \theta + y_p \sin \theta$$
$$H(v, \theta) = H(v, \theta) + 1$$
$$\theta = \theta + 1$$

end

where $\theta_{min}$ and $\theta_{max}$ are the least and highest possible values for the kernel angle respectively. This gives us the accumulator array containing the Hough transform for all edge map points. The angle of the element having the maximum count in the accumulator array is the kernel angle for the channel. The average of the kernel angles for the three channels gives the angle of the motion kernel.

*B. Length Estimation*

The length of the motion kernel is found using a robust cepstral method. Since we now know the direction of blurring, the blurred image can be rotated to make that direction horizontal. Uniform blur produces periodic patterns in frequency domain due to many zero crossings of the sinc function [14]. Figure 4 illustrates this notion.

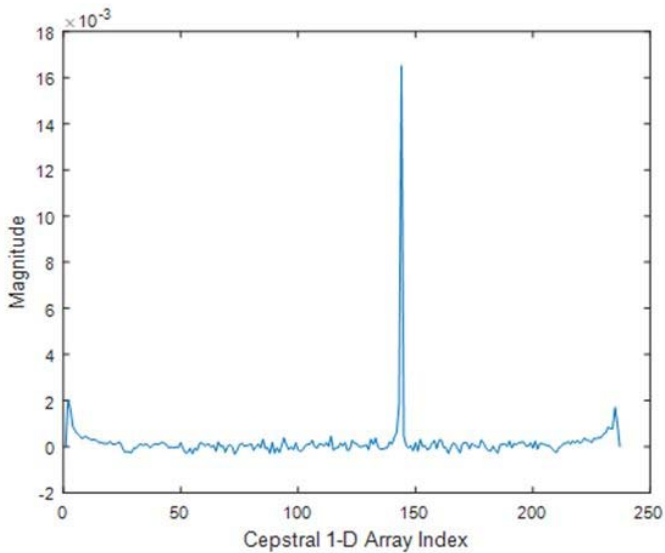

Fig.4. Blurred image cepstrum with a ground truth of $\theta = 70°$, $l = 24$ pixels

We use this property of uniform blur to predict periodicity in the cepstrum of the blurred image which can give us the length of the kernel. First, we compute the cepstrum of the blurred image. This is done using equation 4. The cepstrum is rotated by the kernel angle estimated earlier in the anti-clockwise sense to make the direction horizontal. This rotated cepstrum is transformed into a row vector (1-D array) by taking the mean of the elements in each column.

Looking for the index of the least element in the array from index zero to R, where R is the maximum possible kernel length, gives us the blur length for the channel. The maximum among the lengths found for the three channels gives the length of the motion kernel. Figure 5 shows how the least element index is used in order to find the motion kernel length.

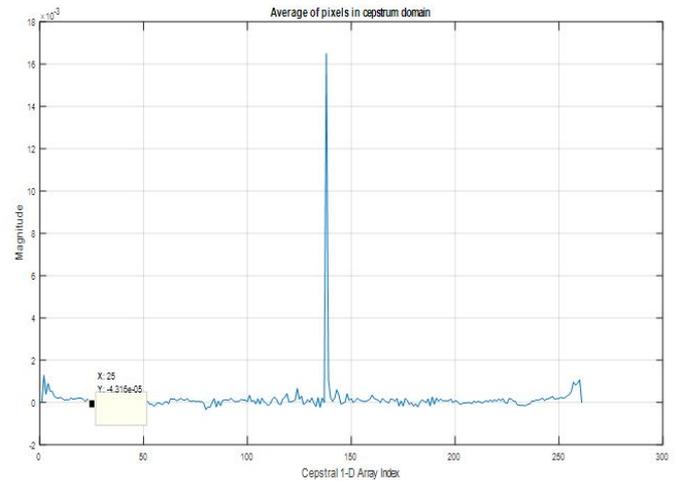

Fig.5. Finding the kernel length for a blurred image with a ground truth: $\theta = 70°$, $l = 25$ pixels

*C. NBID Algorithm*

When we have determined both the parameters that describe the point spread function (PSF), we can create an optical transfer function (OTF) using this PSF. The zeros in the OTF are replaced by small values to avoid the case of division by zero.

We then divide the Fourier transform of our blurred image by this OTF. The inverse Fourier transform of the previous result gives us our de-blurred image.

III. RESULTS AND ANALYSIS

In this section, we talk about the data used in our paper. We will compare the results of our scheme with that of other recently introduced BID methods. Lastly, we analyse the observed results and assess our method's ability to get a good de-blurred image.

*A. Data used in our Paper*

We have used 100 license plate images to test our proposed method. The kernel angle involved in license plate blurring in case of over-speeding usually lies in the range of 40° - 140° [1]. The maximum length of the motion kernel may be in the range of 40 - 50 pixels. The blur angle is determined by a method employing the Hough transform. The blur length is estimated using this angle and the cepstrum of the blurred image. These two parameters are used in the NBID algorithm to get the final de-blurred image.

*B. Evaluation of Performance*

The two stages of kernel angle estimation and kernel length estimation are performed on each blurred image channel. In the angle estimation stage, the mean of the values is taken as the kernel angle while in the length estimation stage, the minimum among the values obtained is taken as the length of the motion kernel.

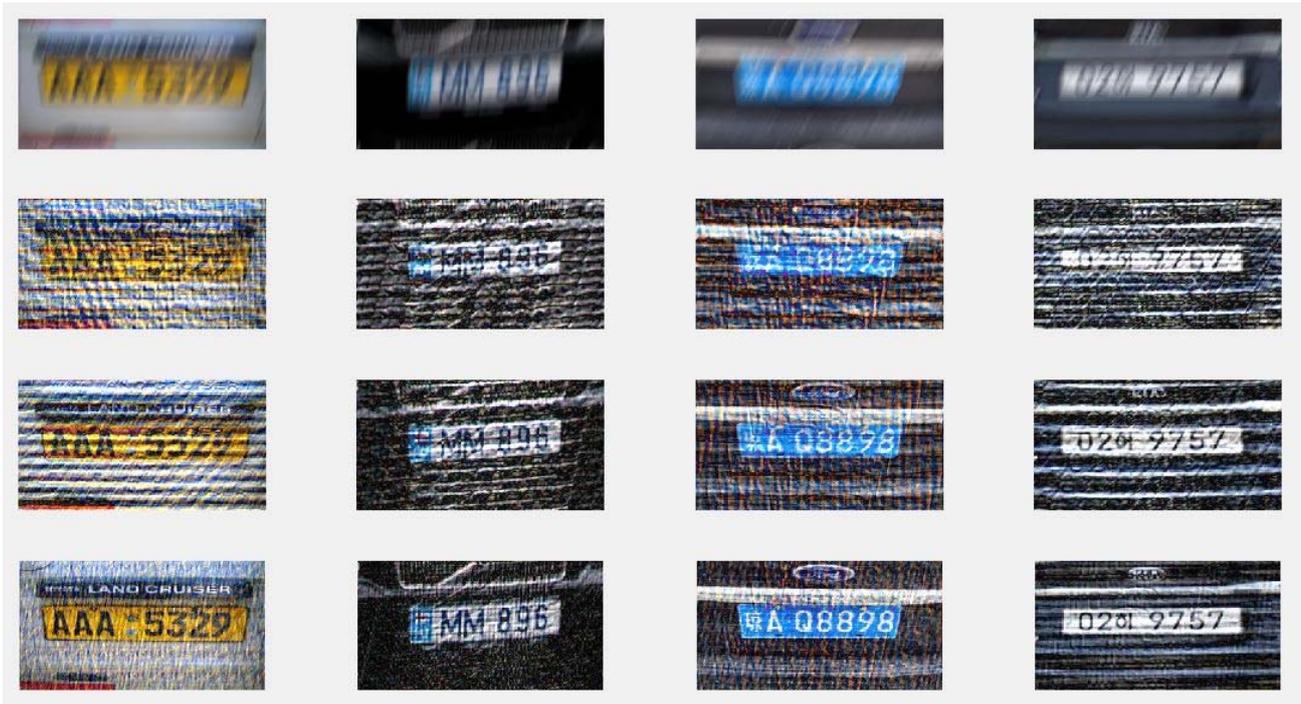

Fig.6. Observed performances of BID schemes: the blurred images are shown in the first row, and the de-blurred results obtained using NSBD [20], RBKE [1] and our method are shown in the second, third and fourth rows respectively.

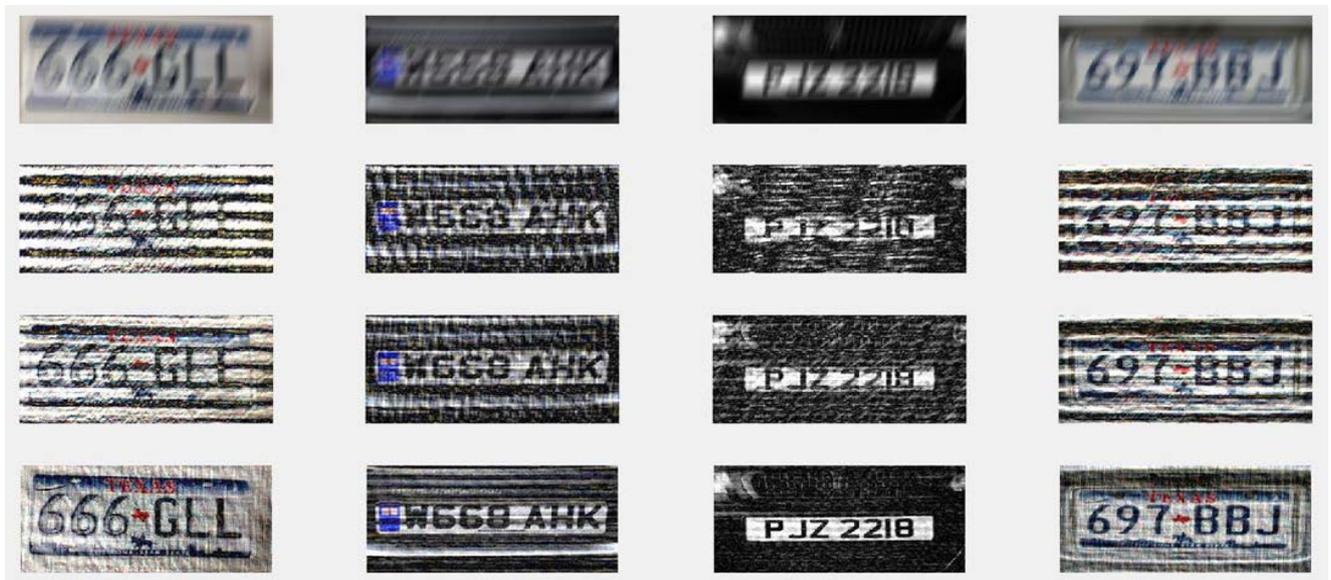

Fig.7. More examples of performances BID schemes: the blurred images are shown in the first row, and the de-blurred results obtained using NSBD [20], RBKE [1] and our method are shown in the second, third and fourth rows respectively.

Our method's performance is compared to that of two recent BID techniques: 1) NSBD [20] and 2) RBKE [1]. The input images are blurred with kernels of various angles and lengths (ground truth). Noise is also added to these images to make the situation real-world. These images are processed using 1) NSBD, 2) RBKE and 3) our proposed scheme. Our aim is to estimate the kernel as accurately as possible to the ground truth. The de-blurred results obtained using these three methods are compared in Figures 6 and 7. For most of the cases, our proposed scheme gives a better final de-blurred output and is shown to significantly improve image quality to make the characters in the license plate more recognizable.

The character recognition rate of our method is also compared to that of other BID schemes. All sharp characters are resized to a fixed size. A support vector machine having a radial basis function kernel is trained to classify the characters as recognizable or not. We set the SVM parameters according to *LIBSVM* [21]. We then apply the pre-trained SVM to 20 blurred images. The various BID methods are compared with their recognition rates in Table I. Clearly, the recognition rate shown by our scheme is an improvement compared to the other schemes.

TABLE I: Comparison of Recognition Rates

| Method | Recognition Rate |
|---|---|
| Blurred Images | 9.1% |
| NSBD [20] | 19.5% |
| RBKE [1] | 72.9% |
| Our method | 82.8% |

The running time of our method is compared with that of recent BID methods. The comparisons are shown in Table II. We can observe clearly that our scheme has a lesser running time compared to these schemes.

TABLE II: Comparison of Running Times

| Method | Recognition Rate |
|---|---|
| NSBD [20] | 17.31s |
| RBKE [1] | 349.81s |
| Our method | 10.49s |

*C. Discussion*

When we see the problem of kernel estimation as one of parametric estimation, we simplify the de-blurring process for the case of blurred vehicle snapshots. The aim is to recover almost all characters in the license plate image so that information about the vehicle can be acquired. The Hough transform gives us good kernel angle estimation, while our new cepstral method gives an accurate estimate of the kernel length. The estimated parameters are quite close to the ground truth. The method adopts a balanced approach and gives an improved de-blurred result.

IV. CONCLUSION

We proposed a new parametric de-blurring method to accurately determine the kernel involved in blurred license plate images. We did this with the aim of handling significant blur in images captured by traffic surveillance cameras to give better quality de-blurred images which can help in recognition of the license plate.

For estimating the kernel angle, we use a method involving the Hough transformation while for the kernel length, we introduce a robust cepstral method to accurately estimate kernel length. When we find these parameters, we can determine the linear motion kernel responsible for causing the blur and de-convolute our blurred image using a simple NBID algorithm.

Our method shows good performance in handling largely blurred vehicle snapshots captured by traffic cameras when compared to recent BID schemes and can be used in traffic surveillance systems to identify troublemaker vehicles.